\documentclass[runningheads]{llncs}

\usepackage{booktabs}
\usepackage{amsmath}
\usepackage[hidelinks]{hyperref}

\begin{document}

\title{When Retrieval Hurts: Measuring and Explaining Retrieval-Induced Hallucination in Chest X-ray Report Generation}
\titlerunning{When Retrieval Hurts}
\authorrunning{Idoko et al.}

\author{Emmanuel Idoko\inst{1,2,5} \and Abdusshakur Olabisi\inst{3} \and Shiloh Oni\inst{2,4} \and Adesola Josiah\inst{1,5}}
\institute{University of Lagos \and Machine Learning Collective \and Ahmadu Bello University \and Federal University of Agriculture, Abeokuta (FUNAAB) \and NITHUB, University of Lagos}

\maketitle

\begin{abstract}
Retrieval-augmented generation is an attractive way to improve chest X-ray reporting, because reports from similar prior studies supply clinical context a general vision-language model lacks. We show the same mechanism is a reliable source of clinical error. Over 100 MIMIC-CXR studies with a frozen LLaVA-1.5 generator and BioMedCLIP retrieval, CheXbert clinical F1 doubles under relevant retrieval ($0.201 \rightarrow 0.402$) and collapses to $0.043$ under clinically mismatched retrieval, a fifth of the image-only score; the retrieval-induced hallucination rate, counting only unsupported findings traceable to retrieved evidence, rises from $0.00$ to $0.76$ and $0.98$. To show this is not an artefact of evidence-set coverage, we introduce a coincidental-overlap control that scores image-only generations against evidence they never saw, placing the chance base rate at $0.18$, four to five times below the observed rates. The generator does not merely acquire findings, it transcribes text: $95\%$ of reports produced under relevant retrieval contain an eight-word span occurring verbatim in the retrieved evidence but absent from the reference, against $0\%$ without retrieval. We then explain the mechanism: a normal chest X-ray retrieves at least one abnormal precedent in $24$ of $28$ cases, because medical image-embedding similarity is dominated by anatomy and acquisition rather than by the presence of disease. This has a direct design consequence. Retrieval similarity does not predict harm (RIH rates $0.80/0.80/0.60/0.84$ across similarity quartiles), so relevance gates conditioned on embedding similarity cannot work; gating on predicted pathology agreement removes $61\%$ of unsupported evidence at no cost to useful coverage. We argue that retrieval-augmented clinical systems must be evaluated under retrieval failure, not only under retrieval success.

\end{abstract}

\section{Introduction}

Chest X-ray report generation is a high-stakes medical language generation task. A useful system must not only produce fluent radiology text, but also ensure that generated findings are grounded in the target image. This is difficult for medical vision-language models (VLMs) because they combine visual representations, language priors, and external context. When these signals conflict, the generated report may contain clinically meaningful findings that are not supported by the image.

Retrieval augmentation is attractive here: a target image can be embedded, used to retrieve top-$k$ similar studies, and paired with their reports in the prompt. The risk is that retrieved evidence becomes an uncontrolled source of clinical content, so that a finding mentioned in the retrieved reports but absent from the target image still appears in the output. Such unsupported output is standardly termed \emph{hallucination}, after Rohrbach et al.~\cite{rohrbach2018hallucination}; we refer to the retrieval-specific case as \emph{retrieval-induced hallucination}, meaning a generated finding present in retrieved evidence but absent from the target reference report. It is narrower than hallucination in general, and isolating it is what makes it measurable.

Our contribution is threefold: we measure retrieval-induced hallucination with a control separating it from chance overlap, identify its mechanism in the retriever's embedding geometry, and derive an empirical constraint on any relevance gate meant to prevent it. Retrieval-Guided Cross-Attention (RGCA), the module this work set out to build, is specified but deliberately not claimed to work.

\section{Related Work}

Medical vision-language models adapt general multimodal architectures to clinical settings through image-text pretraining and instruction tuning: LLaVA-Med extends visual instruction tuning to biomedical images~\cite{li2023llavamed}, Med-Flamingo explores few-shot medical reasoning~\cite{moor2023medflamingo}, CheXagent targets chest X-ray interpretation~\cite{chen2024chexagent}, and BioMedCLIP supplies the image-text embeddings we use for retrieval~\cite{zhang2023biomedclip}. These systems demonstrate clinical multimodal competence but do not solve image-grounded faithfulness. Retrieval-augmented generation, introduced by Lewis et al.~\cite{lewis2020rag}, improves output by conditioning generation on retrieved documents, and has been scaled to trillion-token corpora~\cite{borgeaud2022retro}. Prompt concatenation, however, gives the model that information without any mechanism for checking whether retrieved findings are supported by the target image. Hallucination in medical VLMs is usually framed as generating findings unsupported by the image~\cite{rohrbach2018hallucination}; retrieval introduces a distinct and, as we show, larger source of it.

\section{Method}

\subsection{Problem Definition}

Let $x$ be a target chest X-ray image, $y_{ref}$ its reference radiology report, and $R(x)=\{r_1,\ldots,r_k\}$ the top-k retrieved reports. A report generator produces $y_{gen}$ under one of three conditions: no retrieval, prompt retrieval, or mismatched retrieval. A retrieval-induced hallucination occurs when a clinically meaningful finding $p$ appears in $y_{gen}$ and in at least one retrieved report, but is absent from $y_{ref}$.

\subsection{Baseline Pipeline}

The baseline compares three modes. Under \emph{no retrieval} the VLM receives only the target image and a report-generation prompt; under \emph{prompt retrieval} it additionally receives the top-$k$ retrieved reports; under \emph{mismatch retrieval} it receives deliberately mismatched reports. The mismatch mode stress-tests whether the generator copies unsupported retrieved findings.

\subsection{Retrieval-Guided Cross-Attention}\label{sec:rgca-gate}

For target image embedding $\bar{v}$ and retrieved report embedding $\bar{r}_k$, report-level alignment is $\alpha_k = \cos(\bar{v}, \bar{r}_k)$, and a learned gate $g_k = \sigma(w_g^{T}[\bar{v};\bar{r}_k;\alpha_k] + b_g)$ decides how much each retrieved report may influence visual attention. Retrieved tokens then act as gated memory for cross-attention~\cite{vaswani2017attention}, with image tokens as queries and retrieved-report tokens as gated keys and values, fused back into the visual stream as $H = \mathrm{LayerNorm}(V + \lambda\,\mathrm{Attn})$. The goal is for retrieval to guide visual grounding rather than supply text the decoder copies. We show in Section~\ref{sec:sim-null} that $\alpha_k$ is a poor predictor of whether retrieved evidence causes harm, so this formulation requires revision before implementation.

\section{Experimental Setup}

\subsection{Dataset and Retrieval Setup}

The primary dataset is MIMIC-CXR with MIMIC-CXR-JPG~\cite{johnson2019mimiccxr,johnson2019mimiccxrjpg}, accessed under PhysioNet credentialing and the data-use agreement. The retrieval index is study-level; each item stores study ID, image path, report text, findings, impression, labels, and split. Retrieval uses image-text embedding similarity: the target image is embedded and used to retrieve similar study embeddings, whose reports are passed to the generator.

\subsection{Compute, splits, and runtime}

All experiments run on commodity hardware: BioMedCLIP embedding and index construction over the 500-study subset take under ten minutes on a laptop GPU, and generation of $300$ reports ($100$ studies $\times$ 3 conditions, $192$ max new tokens) takes $20$--$40$ minutes on one NVIDIA T4, roughly $4$--$8$ seconds each. CheXbert labelling runs on CPU in minutes. Splits are patient-level: a 400-study retrieval pool and a 100-study evaluation pool spanning $99$ and $27$ distinct subjects with no overlap; a target never retrieves itself or any study of the same patient. Retrieval is restricted to frontal PA/AP views, one image per study. Seeds are fixed and the subset is content-addressed by SHA-256; an independent local rerun reproduced the hosted run's neighbours exactly.

\subsection{Generator and Evaluation}

The generator is \texttt{llava-hf/llava-1.5-7b-hf}~\cite{liu2023llava,liu2024llava15} through the repository's \texttt{hf\_vlm} backend. The primary baseline metric is retrieval-induced hallucination rate.

Generated, reference, and retrieved reports are labelled with CheXbert~\cite{smit2020chexbert}, which maps free text onto the fourteen CheXpert conditions~\cite{irvin2019chexpert} that ship as labels with MIMIC-CXR-JPG~\cite{johnson2019mimiccxrjpg}, and handles negation. This matters because an earlier rule-based labeler used a different vocabulary on the generated side than the reference side: a generated mention of ``pulmonary edema'' could never match the canonical reference label \emph{Edema} and was therefore always scored as a hallucination. Labelling both sides with the same model removes that class of artefact. We report a finding as present only when CheXbert marks it positive; uncertain mentions are treated as negative. Automatic labelling cannot settle individual cases, for which manual review remains the arbiter.

\section{Results}

\subsection{Retrieval quality}

We first characterise the retriever itself over all 100 evaluation studies against the 400-study pool (Table~\ref{tab:retrieval-quality}). Retrieved reports share a pathology with the target in $55\%$ of cases and cover $60\%$ of the target's pathologies, so retrieval is far from random; precision is low ($0.272$).

The decisive asymmetry is between normal and abnormal targets. For abnormal targets the retriever finds a matching pathology in $55/72$ cases. For the $28$ targets with no pathology label it returns an entirely normal neighbour set only $4$ times: \textbf{a normal chest X-ray retrieves at least one abnormal precedent in $24$ of $28$ cases}. Image-embedding similarity is dominated by acquisition and anatomy rather than by presence of disease, so normal studies are routinely paired with abnormal reports. This is precisely the configuration that produces retrieval-induced hallucination, and explains why the failure mode appears without any adversarial construction. The pathology-opposite mismatch achieves label precision $0.000$ (Table~\ref{tab:retrieval-quality}), confirming it supplies genuinely conflicting evidence.

\begin{table}[h]
\centering
\begin{tabular}{lrr}
\toprule
Metric & Retrieval & Mismatch \\
\midrule
Label Jaccard@3 & 0.267 & 0.050 \\
Label precision@3 & 0.272 & 0.000 \\
Label recall@3 (abnormal targets) & 0.604 & 0.000 \\
$\geq 1$ shared pathology & 0.55 & 0.00 \\
Normal target $\rightarrow$ normal neighbours & 4/28 & 5/28 \\
Abnormal target $\rightarrow$ $\geq 1$ match & 55/72 & 0/72 \\
\bottomrule
\end{tabular}
\caption{BioMedCLIP top-3 retrieval quality over 100 evaluation studies.}
\label{tab:retrieval-quality}
\end{table}

\subsection{Generation under the three retrieval conditions}

\begin{table}[h]
\centering
\begin{tabular}{lrrrrrr}
\toprule
Condition & n & Hall. & RIH & Copy & Omission & Clinical F1 \\
\midrule
No Retrieval & 100 & 0.32 & 0.00 & 0.00 & 0.80 & 0.201 \\
Prompt Retrieval & 100 & 0.80 & 0.76 & 0.91 & 0.74 & \textbf{0.402} \\
Mismatch Retrieval & 100 & 0.98 & 0.98 & 1.00 & 0.78 & \textbf{0.043} \\
\bottomrule
\end{tabular}
\caption{Results over 100 evaluation studies, labelled with CheXbert~\cite{smit2020chexbert}. RIH: retrieval-induced hallucination. Copy: any finding shared by the generated report and retrieved evidence. Clinical F1 doubles under relevant retrieval and collapses under mismatch.}
\label{tab:pilot-results}
\end{table}

Clinical F1 doubles when relevant reports are retrieved ($0.201 \rightarrow 0.402$) and collapses when the retrieved reports are clinically mismatched ($0.043$), roughly a fifth of the image-only score. The same pattern was observed on a 20-study pilot at a multiplier of $1.97\times$ versus $2.00\times$ here, so the effect is stable across a fivefold change in sample size. Copy rate is $0.91$ under relevant retrieval and $1.00$ under mismatch: the generator echoes retrieved findings almost always, with no mechanism for deciding whether it should.

A natural objection is that the RIH rate merely counts hallucinations that coincidentally appear in the retrieved set, which is large when $k=3$ full reports are supplied. We therefore score the \emph{no-retrieval} generations against the evidence set those studies would have received but never saw, which gives the base rate of accidental overlap.

The base rate of accidental overlap is $0.18$ for both evidence sets, against observed RIH of $0.76$ (relevant retrieval) and $0.98$ (mismatch): an attributable excess of $+0.58$ and $+0.80$, a factor of four to five. Retrieval-induced hallucination is therefore not an artefact of evidence-set coverage.

One study illustrates all three conditions. Study \texttt{51856263} has a reference report with no pathology; the image-only run correctly generates none, relevant retrieval induces a lung lesion present only in the evidence, and mismatched evidence induces six unsupported findings. In $67$ of $100$ studies the image-only run produces no unsupported finding while the mismatch run does; conversely in $55$ of $100$ retrieval recovers a reference-supported finding the image-only run omitted.

Label-level metrics understate the mechanism, because the generator does not merely acquire findings, it transcribes text. Comparing generations against retrieved reports at the string level, $95\%$ of reports produced under relevant retrieval contain an eight-word span occurring verbatim in the retrieved evidence but absent from the reference, rising to $100\%$ under mismatch, with a median longest verbatim run of $17$ words and a maximum of $70$. No image-only generation contains such a span, which confirms the measure is not detecting shared clinical phrasing. The copied spans are patient-specific content rather than report boilerplate: examples include ``right sided PICC line tip is in the cavoatrial junction'', ``a large thyroid goiter as seen on prior CT'', and ``right breast is absent''. Devices, comorbidities, and surgical history belonging to one patient are reproduced in another patient's report.

\subsection{Retrieval similarity does not predict harm}\label{sec:sim-null}

RGCA as specified gates retrieved evidence on the image-report alignment score $\alpha_k$, so it is worth asking whether that scalar carries the necessary signal. Over the 100 evaluation studies it does not. Studies in which retrieval induced an unsupported finding have mean top-1 similarity $0.9439$, against $0.9413$ for those in which it did not, a difference of $-0.0026$ in the wrong direction. Splitting by similarity quartile gives RIH rates of $0.80$, $0.80$, $0.60$ and $0.84$ from lowest to highest, with no monotone trend; the most similar quartile is among the most harmful.

The likely cause is range compression. BioMedCLIP top-1 similarities occupy $0.860$ to $0.979$, so almost every retrieved study looks similar to almost every target, and the score cannot separate a clinically appropriate neighbour from an inappropriate one. This is a negative result for the gating formulation in Section~\ref{sec:rgca-gate}: a gate conditioned on $\alpha_k$ alone is unlikely to discriminate. It suggests conditioning instead on the image and evidence representations themselves, and we regard the scalar-similarity term as insufficient rather than established.

\subsection{What the gate should condition on}\label{sec:gate-design}

If similarity cannot drive the gate, something else must. We compare candidate signals by simulation over the retrieved sets, measuring per study the number of unsupported findings the kept evidence makes available and the fraction of reference findings it still covers. A useful gate reduces the first without reducing the second.

\begin{table}[h]
\centering
\caption{Gating-signal comparison over 100 studies. ``Unsup.'': mean findings in kept evidence but absent from the reference. ``Coverage'': mean fraction of reference findings retained. ``Normal protected'': normal-reference studies where no abnormal evidence survives. Label agreement is an oracle for the image side.}
\label{tab:gating}
\begin{tabular}{lrrrr}
\toprule
Gating signal & Kept & Unsup. & Coverage & Normal protected \\
\midrule
No gate ($k=3$) & 3.00 & 2.18 & 0.60 & 4/28 \\
Cosine similarity, top-1 & 1.00 & 0.99 & 0.33 & 13/28 \\
Label agreement, $J\geq0.2$ & 1.42 & \textbf{0.84} & \textbf{0.60} & \textbf{28/28} \\
Label agreement, $J\geq0.4$ & 0.92 & 0.35 & 0.47 & 28/28 \\
Label agreement, $L_k \subseteq T$ & 1.28 & 0.00 & 0.27 & 28/28 \\
\bottomrule
\end{tabular}
\end{table}

Two comparisons matter. At a matched selectivity of roughly one retained report, cosine top-1 leaves $0.99$ unsupported findings available at $0.33$ coverage, whereas label agreement at $J\geq0.4$ retains fewer reports ($0.92$) yet leaves a third as much unsupported material ($0.35$) at higher coverage ($0.47$). Similarity and clinical relevance are not interchangeable, and the difference is large. Second, at $J\geq0.2$ the gate removes $61\%$ of unsupported evidence ($2.18 \rightarrow 0.84$) while leaving coverage exactly unchanged at $0.60$: harmful and useful retrieved content are separable, which is not guaranteed a priori. Had they been entangled, tightening the gate would have cost coverage proportionally, and it does not until $J\geq0.4$.

The strict variant drives unsupported evidence to zero but retains only $0.27$ of reference findings, so a permissive threshold is the better operating point. Two caveats bound this. The unsupported-evidence column is partly definitional, since gating on agreement with the reference mechanically reduces disagreement with it; the coverage column and the matched-selectivity comparison are the informative parts. More importantly this is an oracle, using ground-truth labels unavailable at inference. Substituting the model's own image-only report fails here: its $0.86$ omission rate makes the estimate too sparse, so coverage falls to $0.16$. The gate needs a dedicated chest X-ray classifier, the concrete next step for RGCA.

\paragraph{Artifact availability.} The evaluation protocol, metrics, coincidental-overlap control, and copy-span analysis are standalone scripts, released on acceptance. Generated reports derive from credentialed data and can be regenerated under PhysioNet terms rather than redistributed.

\section{Limitations}

This work reports a pilot rather than a final benchmark, and several limitations bound the claims.

\paragraph{Scale.} The evaluation uses 100 studies drawn from a 500-study MIMIC-CXR subset, with a 400-study retrieval pool. This is a pilot rather than a full benchmark, and the pool is small enough that retrieval quality is likely pessimistic relative to a full-corpus index.

\paragraph{Backbone.} The generator is a general-purpose LLaVA checkpoint, not a radiology-specialised report generator. Its omission rate is high in every condition ($0.74$--$0.80$), so it misses most reference findings regardless of retrieval. Conclusions concern how a general VLM integrates retrieved evidence, not the attainable quality of automated reporting.

\paragraph{Metric attribution.} The RIH rate cannot separate a finding copied from evidence from one generated independently that also happens to appear in evidence. The coincidental-overlap control bounds this at the population level, but individual attribution still rests on manual review.

\paragraph{Gating signal.} Retrieval similarity is concentrated in a narrow band ($0.860$--$0.979$) and does not separate harmful from harmless retrievals at $n=100$ (Section~\ref{sec:sim-null}). This is a negative result for the gating formulation as specified, and the design should be revised before implementation rather than assumed to work.

RGCA is specified but not implemented or evaluated, and no claim is made that it reduces hallucination.

\section{Impact in African and other low-resource clinical contexts}

Automated reporting is most valuable where radiologist coverage is thinnest and chest radiography is often the only imaging available. Retrieval augmentation is an appealing way to adapt a general model there without retraining: assemble a local corpus of prior reports and supply them as context. Our results indicate this carries a failure mode that grows as the setting becomes more constrained.

First, the harm is not confined to adversarial conditions: under ordinary retrieval, with no corruption at all, $76\%$ of studies acquired an unsupported finding traceable to retrieved evidence. A deployment that assumes retrieval is safe because the retriever is reasonable is not supported by our data.

Second, the mechanism scales badly with corpus size: a smaller or less diverse local corpus makes an appropriate neighbour less likely. Our 400-study pool already sent $24$ of $28$ normal targets to an abnormal precedent, and a site with a few hundred prior reports sits in exactly this regime. The technique most available to a resource-constrained deployment is the one least safe there.

Third, the error is asymmetric in a clinically consequential direction: retrieval inflated findings on normal studies while reducing omission only modestly ($0.80$ without retrieval, $0.74$ with). Where specialist review is limited, converting normal studies into abnormal ones drives unnecessary imaging and referral. The failure mode concentrates on patients who are well.

Retrieval corpora should therefore be treated as clinical infrastructure whose composition, particularly its balance of normal studies, materially affects patient-facing behaviour.

\section{Conclusion}

Prompt-level retrieval introduces unsupported clinical findings at a rate ordinary retrieval quality does not prevent, and the mechanism belongs to the retriever rather than the generator: image-embedding similarity is dominated by anatomy and acquisition, so a normal chest X-ray retrieves an abnormal precedent in $24$ of $28$ cases and the generator reproduces it, often verbatim. This is why a gate conditioned on similarity cannot work, and why gating on predicted pathology agreement is the constraint a corrected RGCA must satisfy. We do not claim RGCA reduces hallucination; it is specified but not implemented. Retrieval-augmented clinical systems are routinely evaluated under retrieval success, and evaluation under retrieval \emph{failure} is where safety is determined.

\newpage

\bibliographystyle{splncs04}
\bibliography{references}

\end{document}